\documentclass{article}

\usepackage{arxiv}

\usepackage[utf8]{inputenc} 
\usepackage[T1]{fontenc}    
\usepackage{hyperref}       
\usepackage{url}            
\usepackage{booktabs}       
\usepackage{amsfonts}       
\usepackage{nicefrac}       
\usepackage{microtype}      
\usepackage{lipsum}
\usepackage{graphicx}
\usepackage{mathtools}
\usepackage{amsthm}
\usepackage{bm}
\usepackage{float}
\usepackage{natbib}
\usepackage[capitalize,noabbrev]{cleveref}
\usepackage[textsize=tiny]{todonotes}
\usepackage{lineno}

\title{Habits and goals in synergy: \\
  a variational Bayesian framework for behavior}

\author{
 Dongqi Han \\
  Microsoft Research Asia\\
  \texttt{dongqihan@microsoft.com} \\
   \And
 Kenji Doya \\
  Neural Computation Unit \\
  Okinawa Institute of Science and Technology \\
  \AND
 Dongsheng Li \\
  Microsoft Research Asia\\
  \And
  Jun Tani \\
  Cognitive Neurorobotics Research Unit \\
  Okinawa Institute of Science and Technology \\
  \texttt{jun.tani@oist.jp} \\
}

\begin{document}
\maketitle


\begin{abstract}
How to behave efficiently and flexibly is a central problem for understanding biological agents and creating intelligent embodied AI. It has been well known that behavior can be classified as two types: reward-maximizing \textit{habitual} behavior, which is fast while inflexible; and \textit{goal-directed} behavior, which is flexible while slow. Conventionally, habitual and goal-directed behaviors are considered handled by two distinct systems in the brain. Here, we propose to bridge the gap between the two behaviors, drawing on the principles of variational Bayesian theory. We incorporate both behaviors in one framework by introducing a Bayesian latent variable called ``intention''. The habitual behavior is generated by using prior distribution of intention, which is goal-less; and the goal-directed behavior is generated by the posterior distribution of intention, which is conditioned on the goal.  Building on this idea, we present a novel Bayesian framework for modeling behaviors. Our proposed framework enables skill sharing between the two kinds of behaviors, and by leveraging the idea of predictive coding, it enables an agent to seamlessly generalize from habitual to goal-directed behavior without requiring additional training.
The proposed framework suggests a fresh perspective for cognitive science and embodied AI, highlighting the potential for greater integration between habitual and goal-directed behaviors.
\end{abstract}


\section{Introduction}
\label{chap:intro}


In cognitive science, intelligent agents like humans and mammals, are thought to engage in two types of behavior: habitual and goal-directed \citep{dickinson1994motivational, redgrave2010goal, balleine2010human, dolan2013goals}. Habitual behavior addresses the actions that are performed automatically, without conscious thought or intention, in order to maximize the agent's benefits (rewards), such as seeking for food and avoiding danger. Habitual behavior is typically model-free (MF), meaning that it does not require the agent to consider the detailed consequences of their actions. On the other hand, goal-directed behavior explains the actions that are performed with the aim of achieving a specific goal\footnote{Note that in this paper, \textit{goal} and \textit{reward} are clearly discriminated. \textit{Reward} is a scalar to be maximized, whereas \textit{goal} is the state (e.g., position or visual perception) to achieve.}, such as going to a certain place. Goal-directed behavior is model-based (MB), and it is typically more flexible and responsive to changes in the environment, as it involves conscious decision-making and planning using a environment model \citep{glascher2010states, lee2014neural}.





\begin{figure*}
    \centering
    \includegraphics[width=0.99\textwidth]{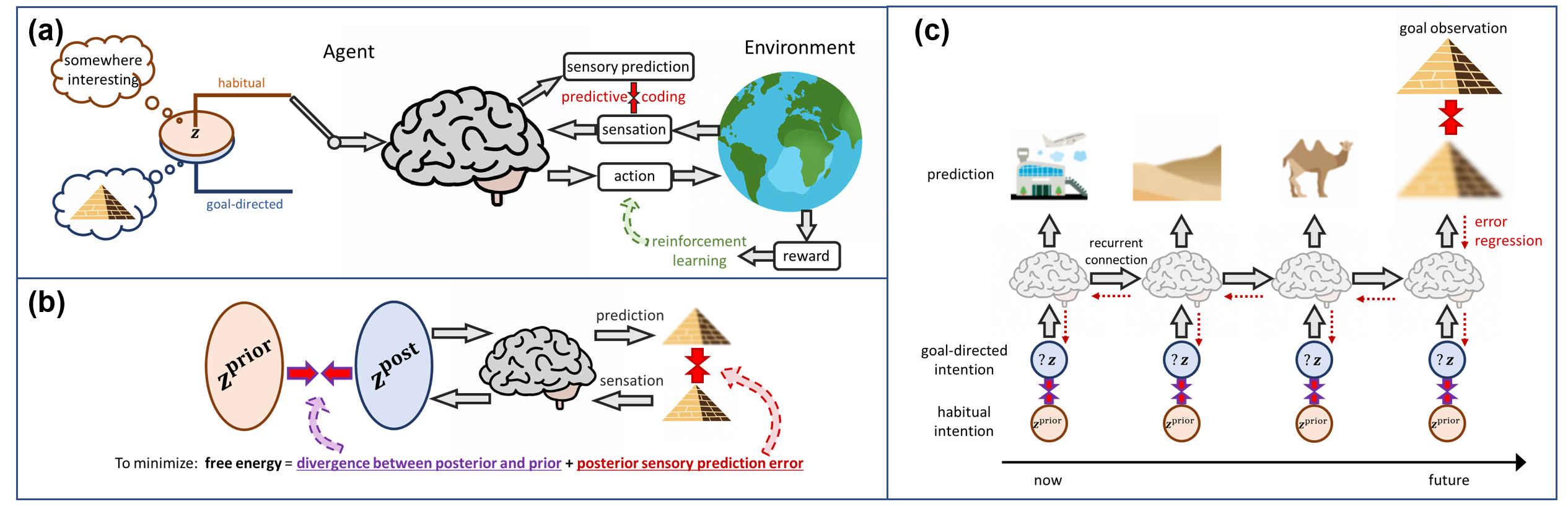}
    \caption[Rough diagrams]{The idea of our Bayesian behavior framework. \textbf{(a)} Behaving using either habitual or goal-directed intention $z$ to generate actions. \textbf{(b)} Minimizing the free energy w.r.t. the desired observation leads to goal-directed intention $z^{\text{post}}$. \textbf{(c)} More details about how to use goal-directed intention by predicting future observations, in which the goal-directed intention ($?z$) is continuously being inferred to minimize the free energy w.r.t. the goal while the prediction model (the ``brain'') is fixed.}
    \label{fig:rough_diagrams}
\end{figure*}

The two types of behaviors have also been extensively studied in machine learning and deep learning, especially in decision-making and control problems \citep{bellman1957markovian}. 
Reinforcement learning (RL) \citep{sutton1998reinforcement} is a computational paradigm that considers to learn a policy (i.e., the strategy to choose actions) that maximize rewards. Model-free RL (MFRL), which does not involve an environmental model, aligns particularly well with acquiring habitual behavior \citep{botvinick2020deep}. On the other hand, the active inference (AIf) theory \citep{friston2010action} appears as a computational framework explaining goal-directed behavior since they both minimized the divergence between the desired goal and the model's prediction conditioned on actions\footnote{It is worth noting that although AIf is also model-based, there are some critical difference between AIf and MBRL (see Section~\ref{chap:related_work:mbrl} and~\ref{chap:related_work:deep_aif}).}.






Conventionally, habitual (MF) and goal-directed (MB) behavior have been treated as two independent problems in both cognitive science and machine learning. Although there are behavioural studies considering a hybrid schema to explain animal or human behavior, the most common way is to simply model the behavior as a linear combination of habitual and goal-directed ones \citep{glascher2010states, smittenaar2013disruption, lee2014neural}. In machine learning, one of practical reason underlies such an separation is that the inputs are different --- In goal-conditioned control or decision making \citep{liu2022goal}, the goal is usually an input to the model. Thus, the model for goal-directed behavior has additional input of the goal compared to that for habitual behavior. Therefore, it is common that two separate models are designed for these two behaviors when both are considered \citep{chebotar2017combining, mendonca2021discovering}.

However, we argue that these two systems should not be isolated from each other. 
Although it has not been fully understood how interactions between these two systems occur in the brain, both habitual and goal-directed behaviors share the same downstream neural pathways such as the brainstem \citep{redgrave2010goal}. The conjecture is that habitual and goal-directed behavior share the low-level motor skills, therefore each system may leverage the well-developed actions learned by the other. Then, how to realize the skill sharing while considering their difference?

In this work, we re-innovate the scheme of behavior from a variational Bayesian \citep{fox2012tutorial} perspective --- we introduce a novel theoretical framework, referred to as \textit{Bayesian behavior}. The proposed framework centers around a probabilistic latent variable $z$ to which we intuitively refer as the ``intention'' of an agent (Figure~\ref{fig:rough_diagrams}). We describe habitual and goal-directed behavior as the prior and posterior distributions\footnote{Prior distributions are probability distributions that represent beliefs about a quantity before having any data, while posterior distributions are probability distributions that represent beliefs about a quantity with data.} of $z$, respectively. In other words, the two behaviors are both drawn from the intention $z$ (and contextual information, such as other brain states), while the difference is that goal-directed behavior is additionally conditioned on the goal than the habitual ones:

\begin{align*}
    \text{habitual action} & \leftarrow z^{\text{prior}} (\text{and contextual information}) \\
    \text{goal-directed action} & \leftarrow z^{\text{post}} (\text{and contextual information})
\end{align*}

where $\leftarrow$ denotes the function (neural pathway) to generate action from $z$ and contextual information. The function is shared for both habitual and goal-directed actions. The prior distribution of $z$ can be any fixed distribution, as it does not depend on the goal. In contrast, the posterior distribution incorporates hindsight information about the agent's future, reflecting the intuitive notion that the current goal-directed action is relevant to a goal to achieve in the future. This additional conditioning differentiates the posterior distribution from the prior distribution, allowing for goal-directed behavior (Figure~\ref{fig:rough_diagrams}a).

In this work, we will demonstrate the aptness of our Bayesian behavior framework to address the following critical questions in cognitive neuroscience by conducting simulated experiments with an embodied robot agent:

\begin{itemize}
    \item [1] How does an agent acquire diverse while effective habitual behavior? 
    \item [2] How to bridge the gap between habitual and goal-directed behavior?
    \item [3] How does an agent generates actions to reach a goal that has not been trained to accomplish?
     
\end{itemize}

We propose that the key neural substrates to address these questions involve predictive coding \citep{huang2011predictive} with the latent Bayesian variable $z$ as a compact representation of current and future sensations. In particular, this is realized by minimizing the variational free energy\footnote{The free energy is mathematically equivalent to the negative variational lower bound in variational Bayesian methods \citep{fox2012tutorial}.} \citep{friston2006free}:

\begin{align}
    \label{eq:free_energy_abstract}
    \text{To minimize:  free energy} = \underbrace{\text{observation prediction error}\left(z^{\text{post}}\right)}_{\mbox{accuracy}} + 
    \underbrace{\text{KL-divergence}\left(q(z^{\text{post}}) \| p(z^{\text{prior}})\right)}_{\mbox{complexity}}
\end{align}

The first term reflects the basic idea of predictive coding by learning an internal model about the environment (Figure~\ref{fig:rough_diagrams}b). The environment model predicts future sensory observations given the agent's intention $z$. A key insight from predictive coding \citep{huang2011predictive} is that $z$ should be much more compact than the original sensory observation by encoding only those information varying with the agent's own intention, as the fixed patterns of the environmental observations should be acquired in the internal model. The compactness of $z$ is crucial for efficient goal-directed planning because it makes plausible an internal searching for proper $z$ that lead to a certain goal (Figure~\ref{fig:rough_diagrams}c).

The second term is the Kullback-Leibler divergence (KL-divergence) \citep{kullback1951information} between the posterior and prior distribution\footnote{In this work, the lower case letter $q$ denotes posterior, and $q(z)$ represents the probability density function of posterior $z$. Similar definition applies to the prior using the symbol $p$.} of $z$, which provides a theoretical foundation to link those two behaviors by bounding their difference (see Section~\ref{chap:elbo_derive} for mathematical explanation). Intuitively, the KL-divergence term balances the fit of the model to the data with the complexity of the latent representation $z$. More discussion can be found in Section~\ref{chap:discussion:pc_ib}.

While predictive coding \citep{huang2011predictive} and the Bayesian brain \citep{doya2007bayesian} have long been discussed in a variety of studies, it has not been used to address the relationship between habitual and goal-directed behavior with detailed examples. Here, we perform proof-of-concept simulated experiments to demonstrate that the neuronal stochasticity does not harm behavior efficiency while enhances diversity. Furthermore, predictive coding enables highly flexible goal-planning capacity of the agent.

The rest of this article is arranged as follows. Section~\ref{chap:background} introduces the basic knowledge about reinforcement learning (RL), predictive coding (PC), free energy principle (FEP) and active inference (AIf) theory. Next, Section~\ref{chap:methods} details the computational methods of the proposed framework, followed by the simulated experiment results in Section~\ref{chap:results}. Then, we briefly clarify how our work relates to and distinguishes from existing methods in machine learning in Section~\ref{chap:related_work}. Finally, we make extensive discussions and conclude this work in Section~\ref{chap:discussion}.

\section{Background}
\label{chap:background}

\subsection{Reinforcement learning}
Reinforcement learning (RL) \citep{sutton1998reinforcement} centers around an agent learning to take actions in an environment to maximize its cumulative rewards. Typically, an RL problem is modeled using a Markov decision process (MDP) \citep{bellman1957markovian}, which is characterized by a set of states, actions, transition probabilities, and rewards. At each time step, the agent takes an action based on its current state and the MDP's transition probabilities, receiving a reward for the action taken. The ultimate objective of reinforcement learning is to identify the optimal policy for the agent to take actions in the MDP that maximize the expected cumulative reward over time. 

There are two main categories of learning methods: model-based (MB) and model-free (MF) \citep{glascher2010states}. Model-based RL involves learning a model of the environment's dynamics and rewards, which is then used to plan actions and compute the optimal policy. Essentially, the agent makes predictions about how the environment will respond to actions and plans accordingly with a computational model \citep{ha2018recurrent}. On the other hand, model-free RL does not involve this explicit modeling of the environment. Instead, the agent learns a direct mapping from states to actions through trial and error, updating its policy based on the rewards received from its interactions with the environment. Popular model-free algorithms include deep Q-network \citep{mnih2015human} and soft actor-critic \citep{haarnoja2018soft}. In both approaches, the agent must balance exploration and exploitation to discover the optimal policy.


\subsection{Predictive coding}
\label{chap:background:predictive_coding}
Predictive coding (PC) \citep{rao1999predictive} is a theoretical framework that suggests the brain employs an internal model, known as a generative model, to generate predictions about incoming sensory information \citep{shipp2016neural}. According to this theory, neural circuits in the brain learn the statistical patterns present in the natural world, reducing unnecessary information by extracting predictable elements from the input and only transmitting what cannot be predicted \cite{huang2011predictive}. In a hierarchical manner, top-down predictions are refined by higher-level cortical areas, while bottom-up processing sends prediction errors upwards to improve the internal model.  The brain uses Bayesian inference to compare its predictions with incoming sensory data, adjusting its internal representation of the world to minimize prediction error.

This process allows the brain to be highly efficient and rapidly adapt to changing environmental conditions. For example, if you have experience with a particular object, your brain will use that experience to generate predictions about what the object should look like, and these predictions will rapidly adjust if the object changes in some way (e.g., if it moves or changes color). By rapidly updating its internal models of the world in this way, the brain can maintain a stable and accurate representation of the environment \citep{ahmadi2019novel, wirkuttis2023turn}.

\subsection{Free energy principle and active inference}

The free energy principle (FEP) \citep{friston2006free, friston2010free} is a Bayesian framework for understanding how biological systems, such as the brain, function. It suggests that biological systems are driven to minimize their free energy, which is a measure of the uncertainty or surprise that the system experiences when it tries to predict the future. The principle is based on the idea that living systems are self-organizing and self-sustaining, and that they use their internal models of the world to make predictions about future events. The free energy principle is related to predictive coding in the same field that states that the brain seeks to minimize prediction error, or the difference between its predictions and incoming sensory data \citep{friston2009predictive, apps2014free}. The free energy principle can be seen as a unifying principle for predictive coding, as it provides a framework for understanding how the brain adapts to its environment and generates adaptive behavior.

Active inference (AIf) \citep{friston2010action, friston2017active} is a form of Bayesian decision-making that is based on the free energy principle. It involves using the brain's internal models to make predictions about the future, and then acting in a way that reduces the prediction error, or the difference between the predicted and actual outcomes. This process allows the brain to actively seek out new information and experiences that reduce uncertainty and increase the precision of its internal models.

\section{Methods}
\label{chap:methods}


We consider the case that the agent first learns habitual behavior and then conduct goal-directed planning (See Sec.~\ref{chap:discussion:first_habit_or_goal} for more discussion about this).

\begin{figure*}
    \centering
    \includegraphics[width=1.0\textwidth]{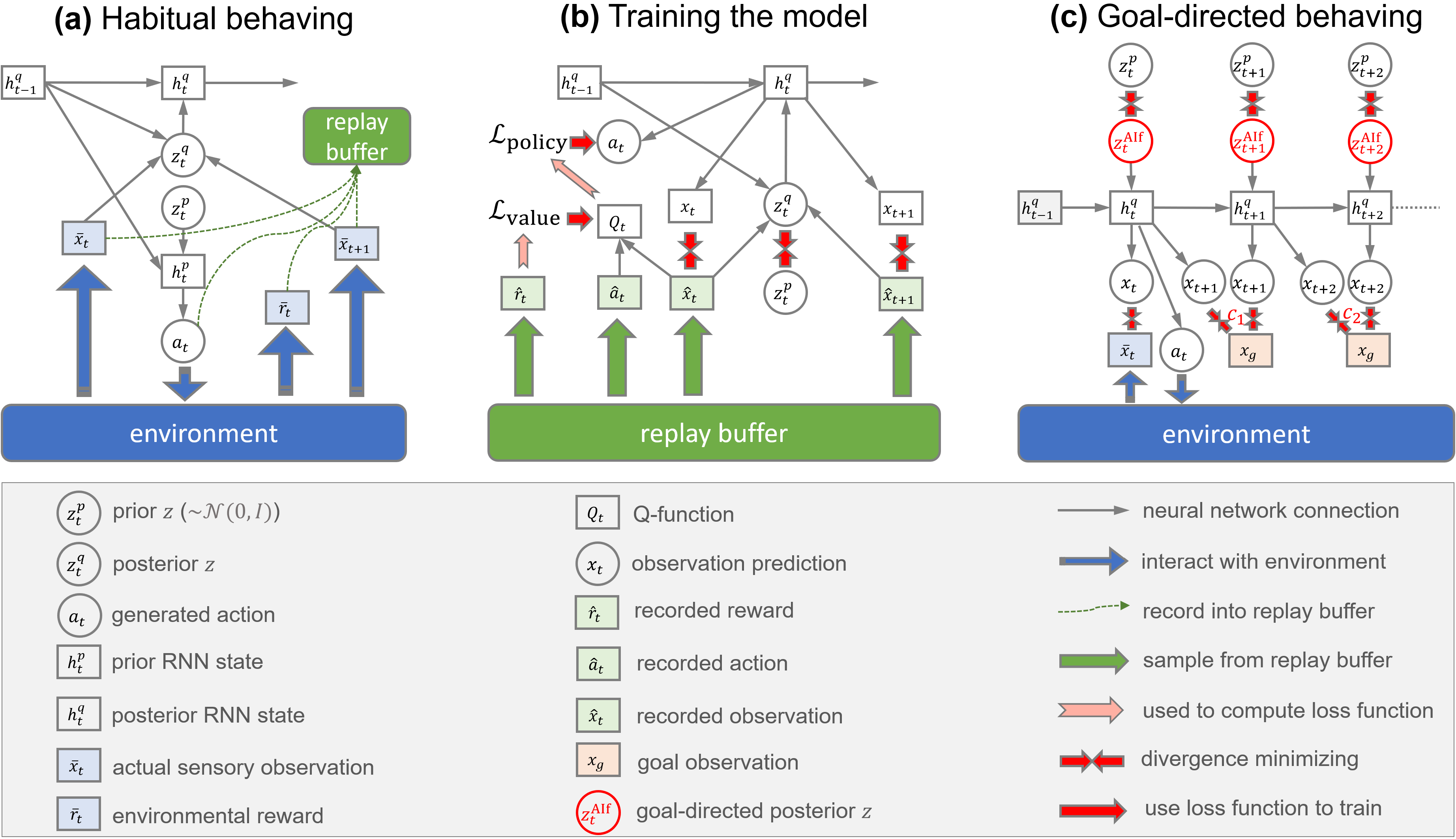}
    \caption[Detailed diagrams]{Detailed workflows of our model in \textbf{(a)} habitual behaving \textbf{(b)} training and \textbf{(c)} goal-directed behaving. See Section~\ref{chap:methods:gdp} for the explanation of $c_1, c_2, \cdots$ in (c).}
    \label{fig:detailed_diagrams}
\end{figure*}

\subsection{Model details}
\label{chap:model_details}

Our model\footnote{The source code is available at https://github.com/FrostHan/BayesianBehavior.} leverage the variational Bayesian method that commonly used in deep learning \citep{kingma2013auto, chung2015recurrent, hafner2018learning}. The core of our model is a 2-dimensional latent variable $z_t$, also referred to as intention in this paper.

A visualized diagram can be checked in Figure~\ref{fig:detailed_diagrams}. The main recurrent neural network (RNN) is a 1-layer gated recurrent unit (GRU) \citep{cho2014learning} which, at step $t$, takes $z_t$ as input and predict the current and subsequence observation ($x_t$ and $x_{t+1}$). We denote the RNN state of the GRU as $h_t$. The decoder which maps $h_t$ to $x_t$ is de-convolutional neural network specified in Table~\ref{table:dcnn_structure}. Another network with the same structure is used for predicting $x_{t+1}$ from $h_t$.  For generating motor actions, a policy network\footnote{As an intuitive understanding, the policy network can be roughly considered as the brainstem which accepts neural signals from both the habitual and goal-directed pathways \citep{dolan2013goals} and outputs motor commands to the body.} is used and trained.

The input to the main RNN, $z_t$, should be paid attention to. $z$ is a Bayesian variable that can be sampled with either its prior or posterior distribution. Correspondingly, we also have prior $h_t^p$ and posterior $h_t^q$ of the RNN state by receiving prior and posterior $z$ respectively. The policy network may take the prior $h_t^p$ as input and give out the habitual action (Figure~\ref{fig:detailed_diagrams}a); or take the posterior $h_t^q$ as input and give out the goal-directed action (Figure~\ref{fig:detailed_diagrams}c). In other word, the habitual and goal-directed behavior share the same policy network. In our implementation, the policy network is a 1-layers GRU followed by a 2-layer multi-layer perceptron (MLP). 

Like in the variational auto-encoder \citep{kingma2013auto}, the prior distribution of $z$ is simply diagonal unit Gaussian $\mathcal{N}(\bm{0}, I)$ since the agent should have no information about the goal in habitual behavior. The posterior distribution of $z$ is also modeled as diagonal Gaussian, of which the mean $\mu^q$ and square-root variance $\sigma^q$ are computed with hindsight information depending on situation, specified in the following sections. Unless specified, the width of each hidden layers is 256 and the activation functions in fully-connected layers are the ReLU function. 

\subsection{Habitual behavior}
\label{chap:methods:habitual_learning}
\subsubsection{Behaving}
We consider a typical episodic\footnote{The terminology ``episode'' is commonly used in RL literature, and is equivalent to the term "trial" used in behavioral studies. In this work, we consider these terms to be interchangeable. Episodic RL deals with scenarios where an agent repeatedly attempts to solve the same task. The agent is reset to its initial state when accomplishing the task or reaching step limitation, and begins a new episode.} RL setting \citep{mnih2015human} for learning habitual behavior, which is a reasonable description of how an animal explores a new environment and learns by trial-and-error. As Figure~\ref{fig:detailed_diagrams}a shows, at environment step $t$, the agent interacts with the environment by computing an action $a_t$ using $h^p_t$. More specifically, the policy network computes a stochastic policy parameterized by $\mu^a_t$ and $\sigma^a_t$, and $a_t$ is given by $a_t = \tanh{(\mu^a_t + \epsilon_t \circ \sigma^a_t)}$, where $\circ$ is the Hadamard (element-wise) production and $\epsilon_t$ follows diagonal unit Gaussian distribution. For better exploration, $\epsilon_t$ is given by pink noise as suggested by \citet{eberhard2023pink}. Then, the agent precepts new observation $\bar{x}_{t+1}$ after executing the action $a_t$ and computes $z^q_t$ to update its RNN state $h^q_t$ (Figure~\ref{fig:detailed_diagrams}a). In particular, $\mu^q_t$ and $\sigma^q_t$ are computed by a 2-layer MLP with hyperbolic tangent activation:

\begin{align}
    \mu^q_t= & \text{MLP}(h^q_{t-1}, \phi(\bar{x}_t; \bar{x}_{t+1})),  \nonumber \\
     \xi^q_t = & \text{MLP}(h^q_{t-1}, \phi(\bar{x}_t; \bar{x}_{t+1})), \nonumber \\
    \sigma^q_t = & \text{softplus}(\xi^q_t) = \ln(1+e^{\xi^q_t}), \label{eq:z_sampling}
\end{align}

where $\phi$ is a CNN (Table~\ref{table:cnn_structure}). The posterior latent variable $z^q_t$ is sampled from the diagonal Gaussian distribution $\mathcal{N}(\mu^q_t, \sigma^q_t)$.
The agent also receive a scalar reward $r_t$ at each step. The environment also provides a termination signal $\text{done}_t$, and the episode (trial) is reset when $\text{done}_t=\text{True}$. 

The agent stores its experience $(\bar{x}_t, \bar{x}_{t+1}, a_t, r_t, \text{done}_t)$ in a replay buffer for experience replay in training \citep{kapturowski2018recurrent} after each step. The replay buffer can store up to $2^{12}$ sequences of length up to 60. The oldest experience will be replaced by a new one when the replay buffer is full.

\subsubsection{Training}


As in typical deep RL with experience replay \citep{mnih2015human}, the neural network models of the agent is updated using stochastic gradient descent each $N$ environment steps, where $N = 10$ in our case. At each update (Figure~\ref{fig:detailed_diagrams}b), a batch (40 sequences of length 60) of experience is randomly sampled from the replay buffer, and all the networks are trained in one gradient step in an end-to-end manner using the following loss function (here using $t$ to denote the step in the recorded sequence, and the loss is averaged over the whole batch):

\begin{align}
    \begin{footnotesize}
    \mathcal{L}  = \underbrace{\underbrace{\mathbb{E}_{q(z)} \left[ \beta_x \ln p(x_t=\hat{x}_t|z_{1:t}) + \ln p_{+1}(x_{t}=\hat{x}_t|z_{1:t-1}) \right]}_{\mbox{posterior prediction error}}  + \underbrace{\beta_z D_{\text{KL}}\left[q(z_t|\hat{x}_{1:t+1}) \| p(z_t)\right]}_{\mbox{complexity}}}_{\mbox{free energy}} + \underbrace{\beta_a \mathbb{E}_{q(z)}\left[\mathcal{L}_{\text{policy}}\right]  + \mathcal{L}_{\text{value}}(\hat{x}_{1:t})}_{\mbox{RL loss function}}, \label{eq:total_loss_habitual}
    \end{footnotesize}
\end{align} 

where $\beta_x, \beta_z, \beta_a$ are the coefficients that determine the balance among these terms.
The term $\mathbb{E}_{q(z)}\left[\mathcal{L}_{\text{policy}}\right]$ is the loss function of policy learning using any RL algorithm conditioned on posterior $z$, where the posterior $z^q_t$ is computed in the same way as in behaving (Equation~\ref{eq:z_sampling}). Note that although the policy loss is expected over the posterior distribution of $z$ in this term, it indeed enhances the performance of habitual actions (using prior distribution of z) if considering the complexity term together (see Section~\ref{chap:elbo_derive}). We here use soft actor-critic (SAC) \citep{haarnoja2018soft, haarnoja2019soft} as the base RL algorithm. In actor-critic algorithms, the value functions, which estimate long-term cumulative rewards over a policy, also need to be learned. We use value networks independent from the main model to learn the Q-function of SAC (the last term $\mathcal{L}_{\text{value}}(\hat{x}_{1:t})$). Each value network is a 1-layers GRU followed by 2-layers MLP. Note that the input to each value network is the original observation encoded by a convolutional neural network (CNN), thus the value network is independent from the main RNN and will only be used in training (Figure~\ref{fig:detailed_diagrams}b) \citep{pinto2018asymmetric}. The hyper-parameters of SAC are selected following \citet{haarnoja2019soft} except that the we change the temperature coefficient to $1.2$ to adapt to our environment. 

The log-likelihood terms in Equation~\ref{eq:total_loss_habitual} $ \mathbb{E}_{q(z)} \left[ \beta_x \ln p(x_t=\hat{x}_t|z_{1:t}) + \ln p_{+1}(x_{t}=\hat{x}_t|z_{1:t-1}) \right]$ are the posterior prediction errors of the current observation $x_t$ and the subsequent one $x_{t+1}$, respectively (Figure~\ref{fig:detailed_diagrams}b). In our work, as image observations are considered, we model each pixel value (ranged in $[0, 1]$) as the probability of a Bernoulli distribution independent from other pixels as in \citet{kingma2013auto}. Therefore, the expectation of the log-likelihood $\mathbb{E}_{q(z)} \ln p(x_t=\hat{x}_t|z_{1:t})$ can be computed as the analytic form:  

\begin{align}
    \mathbb{E}_{q(z)} \ln  p(x_t=\hat{x}_t|z_{1:t}) = \mathbb{E}_{q(z)} [ \hat{x}_t \ln{x_t(z_{1:t})} + (1-\hat{x}_t) \ln{(1-x_t(z_{1:t}))}].  \label{eq:recon_bce_loss}
\end{align}

This also applies to the prediction error for the subsequent observation $\mathbb{E}_{q(z)} \ln p_{+1}(x_t=\hat{x}_t|z_{1:t-1})$. 
The KL-divergence term $D_{\text{KL}}\left[q(z_t|\hat{x}_{1:t+1}) \| p(z_t)\right]$ can be analytically given as both the prior and posterior follow Gaussian distribution \citep{kingma2013auto}. In particularly, as the prior follows $\mathcal{N}(\bm{0}, I)$, the KL-divergence is computed as $D_{\text{KL}}\left[q(z_t) \| p(z_t)\right] = -\ln(\sigma^q_t) + \frac{(\mu^q_t)^2 + (\sigma^q_t)^2}{2} - 0.5$. We choose hyper-parameters $\beta_x=0.1$, $\beta_z=100$ and $\beta_a=30$, obtained by grid search. 

\subsection{Goal-directed behavior}
\label{chap:methods:gdp}
Supposing that the agent has learned diverse habitual behavior (e.g., the agent may move with different routes), our framework allows it to perform \textit{zero-shot} goal-directed planning w.r.t. a given goal. Here, ``zero-shot'' \citep{wang2019survey} means the agent need no additional experiences (than the existing experiences of habitual behavior) to perform goal-directed behavior. That is to say, although the agent has not been trained to accomplish a specific goal in habitual learning, it can generate goal-directed actions if a goal is given by the programmer (How the agent can autonomously select a goal is beyond our scope, see Section~\ref{chap:discussion:where_goal} for more discussion).

Predictive coding and active inference makes this kind of zero-shot goal-directed planning possible, as our recurrent model predicts futures observations with $z_t, z_{t+1}, \cdots$ (Figure~\ref{fig:detailed_diagrams}c, suppose the current step is $t$). Since the intention $z$ is a low-dimensional vector, it is not hard for the agent to infer the goal-directed intentions $(z^{\text{AIf}}_t, z^{\text{AIf}}_{t+1}, \cdots)$ that lead a future toward the goal (Figure~\ref{fig:detailed_diagrams}c), which reflects the idea of active inference. More specifically, we fixed the model weights and bias while treat $((\mu^{\text{AIf}}_t, \sigma^{\text{AIf}}_{t}), (\mu^{\text{AIf}}_{t+1}, \sigma^{\text{AIf}}_{t+1}) , \cdots)$ as trainable variables\footnote{In practice, $\xi^{\text{AIf}}$ is the variable to be trained instead of $\sigma^{\text{AIf}}$ like in Equation~\ref{eq:z_sampling}.}, and optimize them to minimize the variational free energy loss function w.r.t. the goal (the current step is denoted by $t$ and the actual current observation is $\bar{x}_t$):

\begin{align}
    \mathcal{L}^{\text{AIf}}  =  \sum_{\tau=1}^{N} &  \underbrace{c_\tau \left( \ln p_{+1}(x_{t+\tau}=x_g|z^{\text{AIf}}_{t:t+\tau - 1},h^q_{t-1}) + \beta_x  \ln p(x_{t+\tau}=x_g|z^{\text{AIf}}_{t:t+\tau},h^q_{t-1}) \right)}_{\mbox{prediction error of goal obs.}}  \nonumber  \\
    &  + \underbrace{\beta_x \ln p(x_{t}=\bar{x}_t|z^{\text{AIf}}_{t},h^q_{t-1})}_{\mbox{prediction error of current obs.}} + \sum_{\tau=1}^{N}\underbrace{\beta_z D_{\text{KL}}\left[q(z^{\text{AIf}}_{t+\tau} | z^{\text{AIf}}_{t:t+\tau-1}, h^q_{t-1}) \| p(z_{t+\tau})\right] }_{\mbox{complexity constrain}}  , 
    \label{eq:aif_loss}
\end{align}

where the planning horizon $N=16$ in our implementation. Since it is unknown that how many steps does the agent need to take to reach the goal, there are also trainable parameters $c_1, \cdots, c_N$, where $c_\tau$ is a real number in $[0, 1]$ denoting the probability of reaching the goal after $\tau$ steps from now (step $t$), and we have $\sum_{\tau=1}^N c_\tau = 1$. In practice, we optimize real numbers $\varsigma_1, \cdots, \varsigma_N$, and $c_\tau= e^{\varsigma_\tau} / \sum_{i=1}^N e^{\varsigma_i}$ (softmax). Note that since the current observation is known, and it should be used to constrain $z^{\text{AIf}}_t$. One may have found that this loss function does not involve RL loss comparing to that during habitual learning (Equation~\ref{eq:total_loss_habitual}). This is reasonable by assuming the goal is possible to be achieved by habitual behavior, and thus there is no need to re-train the existing motor skills. In practice, we use a batch (32) of planning sequences and optimize them in parallel, using an RMSProp optimizer \citep{hinton2012rmsprop} with decay rate 0.9 and learning rate 0.3. The planning sequence in the batch with the lowest loss after 100 optimization steps will be used. To avoid the random bias brought by sampling $z^{\text{AIf}}_t$ from $(\mu^{\text{AIf}}_t,\sigma^{\text{AIf}}_t)$, we use the mean $\mu^{\text{AIf}}_t$ to approximate $z^q_t$ in rolling the RNN and in the predictive error terms, i.e., the practical AIf loss function is given by

\begin{align}
    \mathcal{L}^{\text{AIf}}  \approx  \sum_{\tau=1}^{N} &  c_\tau \left( \ln p_{+1}(x_{t+\tau}=x_g|\mu^{\text{AIf}}_{t:t+\tau - 1},h^q_{t-1}) + \beta_x  \ln p(x_{t+\tau}=x_g|\mu^{\text{AIf}}_{t:t+\tau},h^q_{t-1}) \right)  \nonumber  \\
    &  + \beta_x \ln p(x_{t}=\bar{x}_t|\mu^{\text{AIf}}_{t},h^q_{t-1}) + \sum_{\tau=1}^{N}\beta_z D_{\text{KL}}\left[q(z^{\text{AIf}}_{t+\tau} | \mu^{\text{AIf}}_{t:t+\tau-1}, h^q_{t-1}) \| p(z_{t+\tau})\right]  . 
     \label{eq:aif_loss_practical}
\end{align}


After the optimization steps explained above, we have obtained the goal-directed intention $z^{\text{AIf}}_t$ with lowest free energy w.r.t. the goal from the training results. Then the RNN state $h^q_t$ is computed as $h^q_t = \text{GRU}(h^q_{t-1}, z^{\text{AIf}}_t)$, and the action is given by $a_t=\tanh(\mu^a_t(h^q_t))$. The active inference process is conducted at each environment step, except that the beginning step of each episode ($t=1$) is the same as in habitual behavior for warming-up.

Notably, the goal here can be of highly flexible choice. The most basic case is that the agent is provided with its visual image as the goal. In this case, the observation prediction error term can be computed in the same way as training (Equation~\ref{eq:recon_bce_loss}). The goal can also be a part of the image by masking the other parts in the prediction error. Furthermore, the goal may also be a specific color, and the prediction error is computed as the difference between predictive future observation and an image full of such a color. In practice, suppose the goal color is $G$ (RGB-value), we replace the prediction error (for both $p$ and $p_{+1}$ in Equation~\ref{eq:aif_loss_practical}) with $\sum_{i,j,c} \exp[-(x_{i,j,c}^{\text{pred}} - G_c)^2/0.5]$, where $x_{i,j,c}$ is the predicted pixel value at the the \textit{i}th row, \textit{j}the column and color channel $c$. Inversely, we may minimize the negative of such a different so that the goal is to observe less of this color by replacing the prediction error with $\sum_{i,j,c} \exp[-(x_{i,j,c}^{\text{pred}} - G_c)^2/0.005]/10$. In these two cases, since the goal is about all future steps, we set $c_\tau=1$ for all future steps. In sum, any goal that can be reflected by a loss function about the observation prediction can be used.

\subsection{On the complexity term in learning habitual behaviors}
\label{chap:elbo_derive}

Here we mathematically show that the KL-divergence between posterior and prior $z$ bridges the gap between the actions based on posterior and prior distributions of $z$. Consider that we want to maximize the logarithm of the expected likelihood of the action computed from prior $z^p_t$ is equal to the optimal action $a^*_t$ (assuming the optimal action is know or can be estimated using the learned value function). The loss function can be written as:

\begin{align}
    \mathcal{L}_{\text{habitual behavior policy}} & := - \ln \mathbb{E}_{p(z_t)}  \left[ P\left(a(z_t) = a^*_t   \right) \right] \nonumber \\ 
    & = - \ln{ \int_{z_t} P\left(a = a^*_t | z_t \right) p(z_t)} dz_t \nonumber \\
    & =  - \ln{\int_{z_t} \frac{q(z_t|\hat{x}_{1:t+1})}{q(z_t|\hat{x}_{1:t+1})} P\left(a = a^*_t | z_t\right) p(z_t) dz_t}. \nonumber
\end{align}

By Jensen's inequality, we have

\begin{align}
     \mathcal{L}_{\text{habitual behavior policy}}    & \leq - \int_{z_t} q(z_t|\hat{x}_{1:t+1}) \ln{\frac{P\left(a_t = a^*_t | z_t \right)p(z_t)}{q(z_t|\hat{x}_{1:t+1})}} dz_t \nonumber \\ 
                    & = - \int_{z_t} \left[q(z_t|\hat{x}_{1:t+1})\ln{P\left(a_t = a^*_t | z_t \right)} - q(z_t|\hat{x}_{1:t+1}) \ln{ \frac{q(z_t|\hat{x}_{1:t+1})}{p(z_t)}}\right]  dz_t \nonumber \\
                    & = \underbrace{- \mathbb{E}_{q(z_t|\hat{x}_{1:t+1})}\left[ \ln{P\left(a_t = a^*_t | z_t \right)} \right]}_{\text{\normalsize posterior policy loss}} + \underbrace{D_{KL} \left[{q(z_t|{\hat{x}_{1:t+1}}})\|p ({z_t})\right]}_{\text{\normalsize complexity}}. \label{eq:vlog}
\end{align}

Thus, minimizing Equation~\ref{eq:vlog} leads to lower $\mathcal{L}_{\text{habitual behavior policy}}$. This derivation is similar to the derivation of variational lower bound \citep{kingma2013auto}. The process can be intuitively understood as the learning with posterior z as oracle information to guide habitual behaviors \citep{han2022variational}.

We can see that the posterior policy loss and complexity terms in Equation~\ref{eq:vlog} are contained in the total loss used in habitual learning (Equation~\ref{eq:total_loss_habitual}), while interestingly, the complexity term  jointly exists with free energy. Equation~\ref{eq:vlog} explains that why Equation~\ref{eq:total_loss_habitual} is actually optimizing the habitual behavior (given by  prior $z^p_t$) despite the policy loss term is expected over posterior $z^q_t$.

\section{Results}

\label{chap:results}

\subsection{Environment}
\begin{figure*}[ht]
    \centering
    \includegraphics[width=0.95\textwidth]{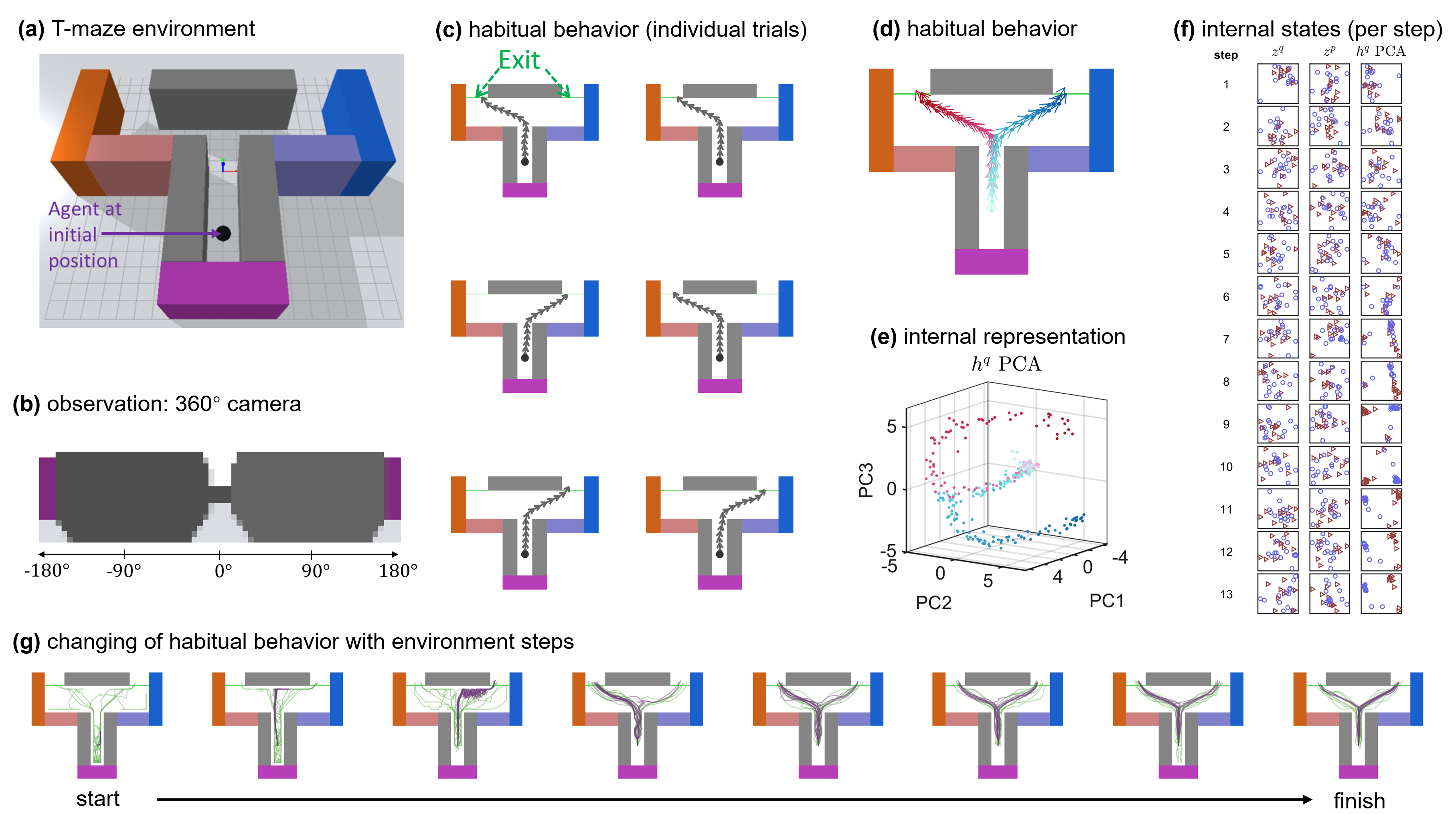}
    \caption[Habitual behaviors in the T-maze environment]{Habitual behaviors in the T-maze environment. \textbf{(a)} rendering of the environment in PyBullet physics engine. The agent is abstracted as a black ball-shaped robot. The agent is rewarded if it escapes from the maze via the top-left or top-right exit. \textbf{(b)} Visual observation of the agent at the initial position, which is the robot's first-person view of the environment using a 360$^\circ$ RGB camera.  \textbf{(c)} Moving trajectories of an example agent's habitual behavior in six different trials, without noise in motor actions (aerial view). The gray square denotes the initial position of the agent and each arrow denotes one step. \textbf{(d)} the example agent's habitual moving trajectories of multiple trials. The color indicates the final escaping exit (red or blue) and step from the start (lightness). \textbf{(e)} Visualizing internal representations of the agent using PCA of the RNN state $h^q$ of the agent. The first 3 PCs are plotted, where data and colors are consistent with (d). \textbf{(f)} visualization of $z^q$, $z^p$, $h^q$ in each step. For $h^q$, PCA is conducted for the data in individual steps, and the first 2 PCs are plotted. The markers correspond to final exit (red triangles: left, blue squares: right). \textbf{(g)} Changing of habitual behaviors from the beginning of learning to convergence. The dark purple and light green curves indicate the trajectories of deterministic and stochastic motor actions in multiple trials, respectively.}
    \label{fig:habitual_result}
\end{figure*}

We focus on a relatively simple yet important navigation environment in a T-shape maze, or simply T-maze (Figure~\ref{fig:habitual_result}a). The T-maze environment is a common behavioral paradigm used in cognitive science to study learning, memory, and decision-making processes in animals \citep{o1971hippocampus, olton1979mazes}. Here we consider a variant of the T-maze, in which the objective of habitual behavior is to escape from the maze as soon as possible, assuming that an enemy is chasing the agent. There are two exits in the top-left and top-right corners (Figure~\ref{fig:habitual_result}a). In each trial (episode), the agent start from a fixed initial position in the bottom (Figure~\ref{fig:habitual_result}a) of the maze. If the agent reaches an exit, it will receive a reward of amount 100 and this trial is finished. Hitting the wall once will bring to a negative reward (-10) to the agent. We consider a discount factor of 0.9 \citep{sutton1998reinforcement} for RL so that escaping with fewer steps is of higher value and encouraged. 

To make the environment more realist like for a biological agent, the observation of the agent is visual perception -- a 360$^\circ$ RGB camera center at the agent with resolution 16 by 64 (Figure~\ref{fig:habitual_result}b). We consider continuous-valued motor action: the agent can decide its horizontal movement at each step (represented by a 2-dimensional action vector) with a speed limit. 


We focus on this environment for two primary reasons. First, the T-maze environment serves as a straightforward and widely adopted paradigm in cognitive science, enabling researchers to investigate various cognitive processes such as spatial learning and memory, working memory, and decision-making \citep{o1971hippocampus, olton1979mazes}. Its simplicity and versatility make it a popular choice for many researchers. Second, we are interested in understanding how an agent can autonomously develop diverse and effective behaviors through trial-and-error. The environment offers a clear illustration of various habitual behavior strategies, such as choosing between the top-left or top-right exits, which requires a decision branching at some point. Since the agent's actions are continuous and it must balance exploration and exploitation trade-offs in a simple yet meaningful context, the autonomous development of this decision branching presents a significant challenge that is under-explored in previous studies. Our framework aims to address this problem with the Bayesian (stochastic) latent intention $z$ which enables the diversity in abstracted action space while keeping the low-level action efficient.

\subsection{Learning diverse and effective habitual behavior}

\label{chap:habitual_results}

The neural network model of the agent is trained by RL and free energy minimization in learning habitual behavior (detailed in Section~\ref{chap:methods:habitual_learning}). After abundant training (400,000 environment steps), the agent acquired diverse (randomly escaping from top-left or top-right) and effective (using few steps without hitting the wall) habitual motor actions. Figure~\ref{fig:habitual_result}c shows the aerial view of one example agent's habitual behavior in six different trials after training, with no randomness in the motor action. The outcome of which exit to escape from is implicitly decided by the neuronal noise of the intention $z$ (Equation~\ref{eq:z_sampling}) in the first several steps of each trial. 

It is also interesting to look into the internal representations of the neural network model. In particular, we visualize the main RNN state $h^q_t$ using its principle components (PC) \citep{pca} based on multiple trials of habitual behavior of the example agent. Figure~\ref{fig:habitual_result}d shows the moving trajectories of the agent in these trials, where red and blue trajectories correspond to the option of escaping from left and right, respectively.  Figure~\ref{fig:habitual_result}e shows the first 3 PCs of $h^q_t$, where the color is consistent with Figure~\ref{fig:habitual_result}d. It can be seen that a branching of the $h^q_t$ representation occurs after the first several steps. This branching is introduced by the randomness of the 2-dimensional latent variable $z$, and we also visualize the $z^q_t$, $z^p_t$, and $h^q_t$ (first 2PCs) in each respective step in Figure~\ref{fig:habitual_result}f, where the color indicate the final escaping exit. An interesting result is that the branching is not happened at a single step, which can be seen from the fact that there is no clear separation of red $z^p$ and blue $z^p$ at any single step (Figure~\ref{fig:habitual_result}f) 
Nonetheless, the branching is fully decided by $z$, which will be verified in Section~\ref{chap:gdp_results}. We also plot the development of agent' habitual behavior in the progress of learning in Figure~\ref{fig:habitual_result}g, which demonstrate the agent gradually develops diverse and efficient habitual behavior using RL (trial and error).

\begin{figure}[t]
    \centering
    \includegraphics[width=1.0\textwidth]{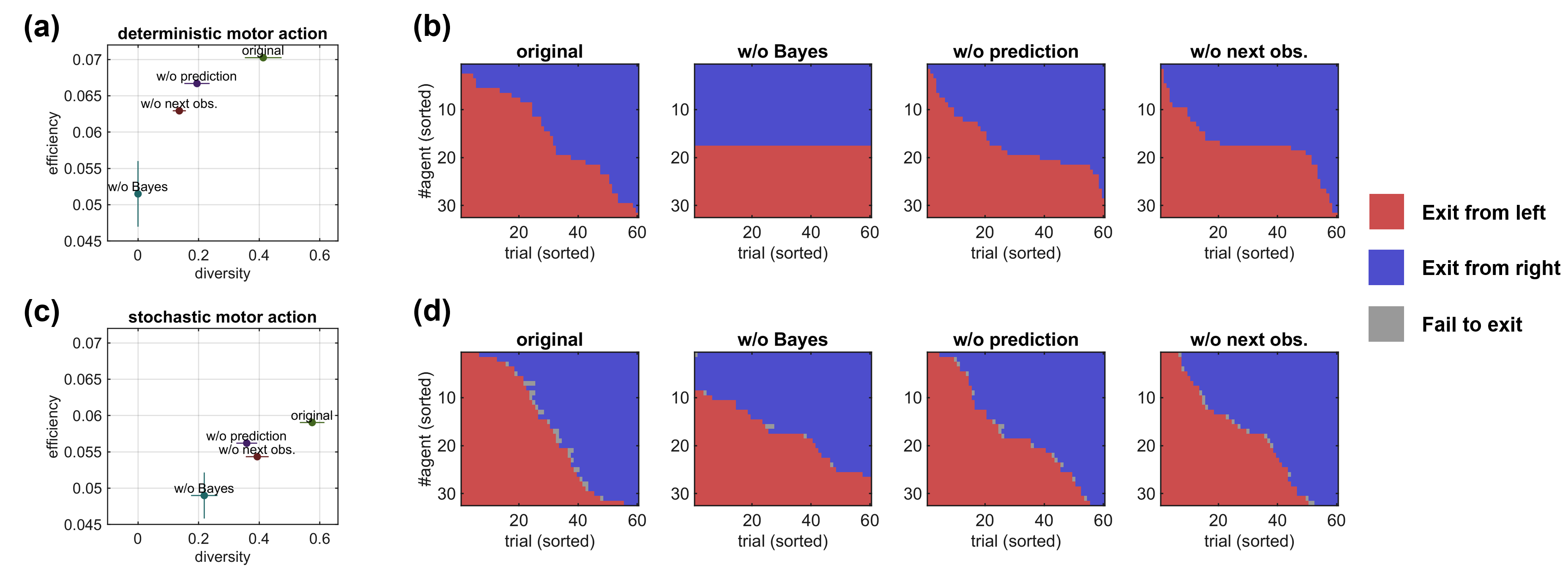}
    \caption{Ablation studies of habitual behavior. \textbf{(a)} Diversity and efficiency. Efficiency is defined as $1/\mbox{mean}(\text{steps to reach exit})$ and diversity is computed as $\mbox{min}({n_\text{left}}/{n_\text{right}}, {n_\text{right}}/{n_\text{left}})$, where $n_\text{left}$ and $n_\text{right}$ are the number of trials successfully exiting from the top-left and top-right corner, respectively, in 60 testing episodes. The models are explains as follows: \textbf{No Bayes}: This model is the same as the original one except that $z$ is a fully deterministic variable, and there is no prior distribution of z thus the complexity term is not involved. \textbf{No prediction}: This model does not perform observation prediction, otherwise it is the same as the original one. \textbf{No next obs.}: In this case, it differs from the original model in terms that the posterior $z^q_t$ does not predict and depend on the subsequent observation $x_{t+1}$. \textbf{(b)} Number of trials exiting from left or right for each agent (random seed). For (a) and (b), 32 random seeds are used where the motor actions are deterministic. \textbf{(c)} and \textbf{(d)} show the results using stochastic motor actions.}
    \label{fig:exit_ratio}
\end{figure}

As our framework involves a number of mechanisms, we conducted ablation study to understand the roles of three crucial parts of our framework learning habitual behaviors. We compute the \textit{diversity} and \textit{efficiency} metrics to quantify the performance (Figure~\ref{fig:exit_ratio}). The results show that these components are crucial for developing more diverse and efficient habitual behavior. 

\subsection{Flexible goal-directed behavior immediately transferred from habitual behavior}

\label{chap:gdp_results}

In habitual learning, the agent has acquired the behavior that randomly going to left or right exit. The experiences in habitual learning also helped the agent to form an internal predictive model of visual observations in the T-maze environment. An intelligent embodied AI should also have the ability to perform the goal-directed behavior without extra training, which is indeed allowed by our framework using the method detailed in Section~\ref{chap:methods:gdp}. 

Conventional goal-conditioned RL \citep{liu2022goal} (Section~\ref{chap:related_work:goal_conditioned_rl}) treat the goal as an input to the model and outputs goal-directed action. Such a treatment has two crucial limitations. First, the goal needs to be explicitly involved in training, in which the agent is rewarded when it achieves the given goal. (Section~\ref{chap:related_work:goal_conditioned_rl}). Second, the full state of the goal needs to be given at each trial. For example, if the desired goal is to ``observe more red colors'', it is unknown how to provide the input. Simply using an all-red image may not be proper if there is no such an state. In contrast, our framework, which is based on PC, considers the goal by using the variational free energy as a loss function to minimize (Equation~\ref{eq:aif_loss}). The free energy w.r.t. the goal is much more flexible. Also consider the case that the desired goal is observing more red colors, we can simply replace the prediction error term in the free energy (Equation~\ref{eq:aif_loss}) by a loss function reflecting how red the predicted image is. 

\begin{figure*}[t]
    \centering
    \includegraphics[width=0.85\textwidth]{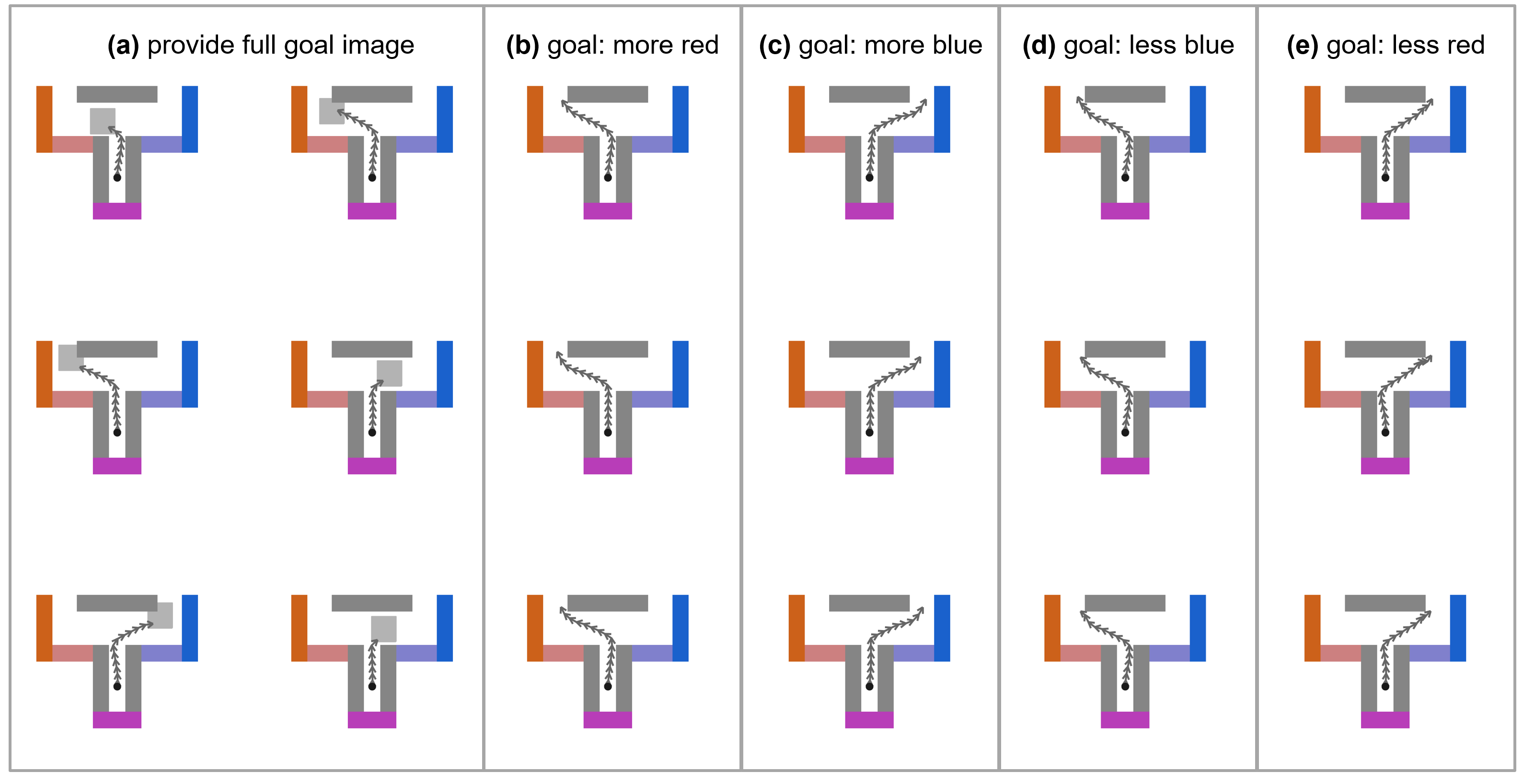}
    \caption{Moving trajectories of the example agent's goal-directed behavior in different trials, without noise in motor actions (aerial view), plotted in the same way as in Figure~\ref{fig:habitual_result}c. The light gray square in \textbf{(a)} indicates the goal area (the goal image is the visual observation in the center of the goal area).}
    \label{fig:planning_trajectories}
\end{figure*}

Furthermore, our framework enables the agent to achieve a goal that is not explicitly involved during RL, i.e., zero-shot transfer from reward-maximizing, habitual control to reward-free, goal-directed planning. The key mechanism to emerge such an ability is a compact representation of behaving strategies using the latent intention $z$ which is also bound to future observations. Here, goal-directed behavior is generated by active inference \citep{friston2016active}: inferring $z$ that minimize the free energy w.r.t. to the goal (Equation~\ref{eq:aif_loss}), i.e., reducing the gap between predicted future outcome and goal while annealing the complexity of $z$ (this $z$ is denoted as $z^{\text{AIf}}$). Then, $z^{\text{AIf}}_t$ can be thus to update the main RNN state\footnote{Here, $z^{\text{AIf}}$ should be considered as the posterior distribution since it is decided by the desired future observation.} $h^q_t$ and compute the action from $h^q_t$ using the policy network. Note that the policy network is the same one used in habitual behavior, which means it tends to output actions that lead to more rewards. In this environment, the actions correspond to reaching any exit faster and avoiding hitting the wall. While going to the left or the right exit from the initial position takes the same amount of efforts, the goal will decide which way to go. Also, the constrain by the KL-divergence between the prior and posterior in the free energy (Equation~\ref{eq:aif_loss}) can be intuitively understood as playing the role to ensure that the inferred $z^{\text{AIf}}$ will not be too much different from the prior distribution (corresponding to the reward-maximizing habitual behavior), so that the agent should not execute inappropriate actions (e.g., hitting the wall).

To demonstrate the flexibility of goal-directed behavior using our framework, we performed experiments with providing three kinds of goals to the agent: (1) going to a place so that the agent's visual observation is the provided image (Figure~\ref{fig:planning_trajectories}a) (2) observing a color as much as possible (Figure~\ref{fig:planning_trajectories}b,c) (3) avoiding observing a color (Figure~\ref{fig:planning_trajectories}d,e). The goal-directed behaving trajectories of the same agent in Figure~\ref{fig:habitual_result}. Each panel shows one kind of goal (detailed methods are explained in Section~\ref{chap:methods:gdp}). And the agents performs goal-directed behavior with high success rate ({a}: 97.9\%, {b}: 97.4\%, {c}: 100\%:, {d}: 99.5\%, {e}: 100\%, tested using 32 agents and 6 episodes for each agent), where a goal-directed trial is considered successfully if the agent reaches the place near the goal in (a), the agent goes to top-left ({b, d}), or the agent goes to top-right in ({c, e}).

\begin{figure*}[t]
    \centering
    \includegraphics[width=0.99\textwidth]{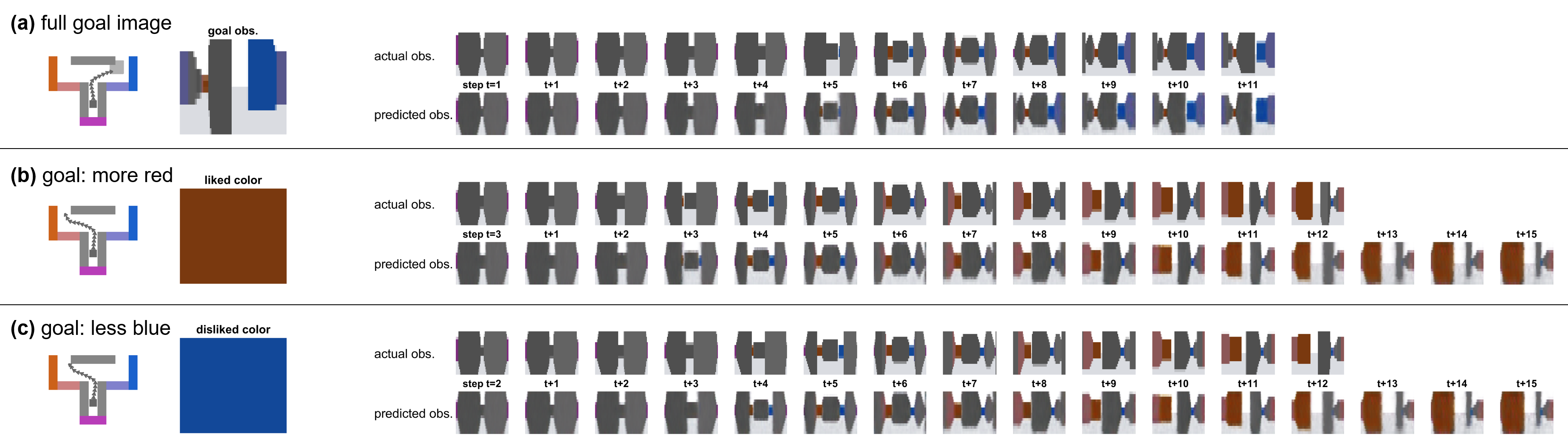}
    \caption{Left: the moving trajectory of the example agent (the same one as in Figure~\ref{fig:planning_trajectories}), plotted in the same way as in Figure~\ref{fig:planning_trajectories}; and the goal (or disliked) observation image (Section~\ref{chap:gdp_results}). Right: the first row shows the post-hoc future observation from the current step $t$. The second row shows the predicted visual observation.} 
    \label{fig:planning_future}
\end{figure*}

Figure~\ref{fig:planning_future} provides more details about the inferred $z^{\text{AIf}}$ for three kind of goals (see Section~\ref{chap:methods:gdp} for detailed methods). First, Figure~\ref{fig:planning_future}a shows the future observations predicted with the inferred goal-directed intention (Section~\ref{chap:methods:gdp}), at the second step in this trial. The agent makes accurate prediction about more than 10 future steps to reach the place where the agent can observe the goal image. It can be also seen that in the case of the goal is to observe more red colors (Figure~\ref{fig:planning_future}b), the agent also makes reasonably realistic future predictions containing more red colors. Figure~\ref{fig:planning_future}c demonstrates another interesting case: the goal is to avoid observing blue colors. The agent also succeed in going to the exit according to this goal.

The predictive ability demonstrated in Figure~\ref{fig:planning_future} is not specific for a few agents, but common in most of the agents. It can be seen that the goal-directed behavior in these cases is approximately covered by the habitual behavior given that the learned habitual behavior contain the diverse options of escaping from either the left exit or the right one. This is one of the most important concepts in our framework -- goal-directed planning is constrained by habits in terms of low-level motor skills. For example, if a person want to move a mug from a place to another, usually the person's habitual gesture to hold the mug will be used instead of other gestures that may also hold the mug. We consider this an essential property for efficient goal-directed planning because the already shaped habits largely reduce the search space of possible actions for the goal\citep{parr1997reinforcement, konidaris2009skill}. Thus, active inference in our framework conduct a search of only ``good'' (reward-maximizing) actions that may lead to future observations close to the goal. Although there is a limitation that the agent may not be able to achieve a goal that is non-reachable by its habitual behavior, this property is crucial to address the efficiency-flexibility dilemma.





\section{Related work in AI}

\label{chap:related_work}

Looking into each part of our framework, many, but not all are well-formulated in literature, or sharing similar ideas. However, our framework is not an incremental study on any existing one of a straightforward combination of them. This section clarifies the novelty of our methodologies (problem define, optimizing algorithm and model architecture) given existing ones. 

\subsection{Embodied artificial general intelligence}
\label{chap:related_work:agi}

Let us first look at position papers about how to create a general, intelligent embodied artificial agent (also known as the foundation model \citep{bommasani2021opportunities}, which refers to AI models that can perform a wide range of tasks and adapt to new challenges). While no work has achieved this goal yet, there are countless articles addressing the problem. Here, we discuss two particularly interesting and related ideas.

The \textit{one big net} framework proposed by Schmidhuber \citep{schmidhuber2018one} stems from the observation that humans and other animals have one large neural network (i.e., their brain) that can efficiently learn and perform a wide range of tasks. Schmidhuber envisions that such a network would be able to continually learn and adapt to new tasks by reusing and transferring knowledge across tasks, making the learning process more efficient. A shared key idea between our framework and the one big net is that the center part of the model is an RNN. The RNN models the physical dynamics of the world, which is intrinsically invariant w.r.t. time, and maintains its internal states which can theoretically encode information from infinitely long history \citep{hochreiter1997long}. However, it is not explicitly pointed out how prediction plays a role in goal-directed planning. 

Another recent perspective from Lecun is the so called \textit{autonomous machine intelligence} framework \citep{lecun2022path}, which shares some common high-level ideas with ours. Lecun emphasizes that the world model, which plays the two-fold role of planning for future and estimating missing observation, should be a energy-based model. In the context of goal-directed behavior, the model takes the current state, the goal and the action to take as input, and outputs a scalar energy value to describe their ``consistency''. Similar to our ideas, the model and internal states are optimized for energy minimization with gradient methods. A key difference is that our model makes explicit predictions about sensation, and the uncertainty of sensation is handled by the stochastic latent variable z with variational Bayesian methods.



\subsection{Model-based RL}
\label{chap:related_work:mbrl}

Model-based RL (MBRL) approaches train the agent(s) based on a mathematical model that can predict upcoming state or observation given current and previous observations and actions. Usually, the model is used either for dreaming, i.e., generating imaginary experiences to train the agent \citep{deisenroth2011pilco, ha2018recurrent, kaiser2019model, hafner2019dream} or planning, i.e., inferring the policy that leads to maximum returns in the future \citep{hafner2018learning, ke2019learning, schrittwieser2019mastering}. Notably, there are also methods that used the model to extract information from the environment to serve model-free RL \citep{igl2018deep, han2020variational, lee2020stochastic}, which share the similar idea with our framework when learning habitual behavior. While \citet{igl2018deep, han2020variational, lee2020stochastic} also used variational RNNs, they focused on single-goal tasks. By contrast, the stochasticity in our model reaches its potential to enable the agent to randomly pursue one of multiple goals. The planning phase of our framework used active inference \citep{friston2010action}, which infers the policy using the model like in MBRL, while not to maximum returns but minimize free energy w.r.t. the goal (see Sec.~\ref{chap:related_work:deep_aif}).

\subsection{Variational Bayes in deep learning and RL}
Variational Bayesian (VB) approaches in deep learning has been popular since the introduction of the variational auto-encoder (VAE) \citep{kingma2013auto, sohn2015learning}. The core idea is to maximizing the variational lower bound of an objective function of a probabilistic variable so that we can replace the original distribution with an variational approximation \citep{alemi2017deep}. \citet{chung2015recurrent} complemented VAE with recurrent connections by proposing variational RNNs. The variational RNN and its variants were later used in deep RL, e.g., as the world model in MBRL \citep{hafner2018learning, hafner2019dream, han2020variational, lee2020stochastic} and for approximating unobservable environmental states \citep{igl2018deep, han2022variational}. One critical reason to use VB models in RL is that they are believed to extract useful representation of environmental state from raw observations \citep{alemi2017deep} and are robust in training. The acquired representation is then incorporated for the original RL task, i.e. maximizing rewards. While our framework can be considered as a new member of the VB family that handles decision-making/control tasks, our idea of modeling habits and goals using variational Bayes has not been discussed by previous deep learning studies.

\subsection{Goal-conditioned RL}
\label{chap:related_work:goal_conditioned_rl}

Goal-conditioned RL (GCRL) \citep{liu2022goal} addresses scenarios in which a goal is given at each episode and needs to be achieved. The goal can be a property or feature \citep{florensa2018automatic}, observation or state \citep{andrychowicz2017hindsight}, or language description \citep{luketina2019survey}. The main difference between GCRL and our framework is that in GCRL, goals are given during learning and agents are only rewarded when the given goal is achieved. In contrast, our framework does not assign any specific goals during training, it is only maximizing rewards for habitual behavior and minimizing free energy expectations. In simple terms, the training and testing problems are consistent in GCRL, but different in the proposed framework.

\subsection{Deep active inference}
\label{chap:related_work:deep_aif}

Active inference (AIf) \citep{friston2016active, friston2017active} has recently been combined with neural networks (a.k.a. deep AIf) to solve more challenging tasks including simulated decision problems \citep{ueltzhoffer2018deep, millidge2020deep, fountas2020deep, mazzaglia2021contrastive} and planning with real robots \citep{ahmadi2019novel, queisser2021emergence, matsumoto2022goal, wirkuttis2023turn}. 

Deep AIf shares the similar idea with MBRL in terms of inferring actions to achieve desired outcome using a environment model, while the main difference is that AIf maximizes the likelihood to achieve a certain state, while MBRL maximizes expected rewards. Another notable difference is that AIf is a probabilistic framework, while MBRL does not have to be. 

Goal-directed planning in our framework employed the idea of active inference. However, our model does not directly infer actions to achieve the goal, but the latent state in the model that encodes the intention. The latent state can be also understood as the high-level action from the perspective of hierarchical RL \citep{sutton1984temporal, eppe2022intelligent}.




\subsection{Control as probabilistic inference}

Control as probabilistic inference (CPI) \citep{levine2018controlAsProbInf} proposes using probabilistic inference to compute the optimal control action instead of designing a deterministic control policy, by casting the control task as a probabilistic inference problem over latent variables that describe the state of the system. Although the basic idea shares insights with model-based learning, CPI does not consider detailed outcomes of action but only maximizes rewards. Therefore, the practical implementation of CPI turned out to be model-free algorithms, such as soft actor-critic (SAC) \cite{haarnoja2018soft, haarnoja2019soft}. In our implementation, SAC is used as the base RL algorithm to learn the habitual behavior. Readers are also encouraged to refer to \citet{millidge2020relationship, hafner2020action} for the in-depth discussion on the relationship between probabilistic inference and decision-making/control.

\subsection{Generalization in RL}

There are also a number of studies done on a problem known as generalization in RL \citep{kirk2021survey}. They consider the cases where training and testing tasks are different in terms of state distributions \citep{cobbe2019quantifying}, dynamics \citep{ni2022recurrent}, observation \citep{cobbe2019quantifying} and reward functions \citep{yu2020meta}. Our framework generalizes habitual control to goal-directed planning. However, as far as we are aware, there has not been any study that handles learning without goals but testing with goals, i.e. our work may be considered as a novel setting of generalization in RL.

\subsection{Self-discovery of goals or skills}

A class of methods known as variational skill discovery (VSD) \citep{achiam2018variational} aims to discover action primitives in reinforcement learning (RL) by optimizing an unsupervised or self-supervised objective function based on information theory. These methods use a latent variable, $z$, to label action primitives, which can be either discrete \citep{gregor2016variational, achiam2018variational, eysenbach2019diversity} or continuous \citep{sharma2020dynamics, xu2020continual}. The policy model, $\pi(a|z,s)$, where $a$ is action and $s$ is state, is then trained using pseudo (intrinsic) rewards and RL. These pseudo rewards reflect an information-theoretic objective that encourages the skills to be diverse and predictable by states, such as the variational lower bound of the mutual information between the set of skills and skill termination states. 

A particularly related work is \citet{mendonca2021discovering}, which, like us, considers both control without and with a goal. However, they explicitly train a goal-achieving agent to achieve goals using RL by designing a ``goal achievement reward'' in addition to training an agent for exploration; While in our framework, there is no training for goal-achieving.

\section{Discussion}
\label{chap:discussion}

\subsection{Computing the Bayesian latent state \textit{z}}

Some readers may still be confused about the Bayesian variable $z$ -- how to compute $z$ in different cases? In this work, the prior of $z$ always follows a diagonal unit Gaussian distribution, which corresponds to diverse and unconditioned choice of goals at current step. In contrast, the posterior of $z$ is aligned with the future states. To be clear, there are two ways to compute the posterior distribution, respectively used in learning habitual behaviors (optimizing the model weights and biases) and goal-directed planning (optimizing $z^{\text{AIf}}$). 

During executing and learning the habitual behaviors (Figure~\ref{fig:detailed_diagrams}a, b), the way to compute the posterior $z^q_t$ is to use the previous RNN state $h^q_{t-1}$, the current observation $\bar{x}_t$ and the post-hoc, next-step observation $\bar{x}_{t+1}$ to as input (a forward process, Equation~\ref{eq:z_sampling}). This is a crucial mechanism to better bind the intention $z$ with the transition of environment state, without a goal being assigned. Since achieving a goal in the future depends on a chain of environment state transitions. Thus, the main RNN using the intention $z$ as input give raises to the capacity for later goal-directed planning. The forward computing way also ensures the computational efficiency in habitual behaviors.

The other way, used in goal-directed behavior (Figure~\ref{fig:detailed_diagrams}c), is active inference, or searching posterior $z^{\text{AIf}}$ that minimizes the free energy w.r.t. the goal (a backward process, Equation~\ref{eq:aif_loss}). Such way of inferring $z^{\text{AIf}}$ is opposite to the conventional algorithms of goal-directed control \citep{liu2022goal} in which the goal is encoded as an input (forward process). One key difference is that active inference allows the training (habitual here) and testing (goal-directed) problems to be different, since it can utilize the learned world model with compact intention representation $z$ to do imaginary future planning (Figure~\ref{fig:detailed_diagrams}c). In contrast, the conventional goal-as-input approach needs to keep assigning the goal in training. Another significant advantage of our approach is that it is flexible to set various properties of observation as the goal, such as seeing more red colors (Section~\ref{chap:gdp_results}), in contrast to the goal-as-input approach which requires a full goal involved in training. Meanwhile, active inference needs much more computations to conduct goal-directed planning (optimizing $z^{\text{AIf}}$), which is unavoidable and also well known in animal behaviors \citep{redgrave2010goal}.


\subsection{Toward understanding behavior}
Recall the three questions raised in Introduction, we now provide our answers to them after demonstrating our experimental results:

\textbf{How does an agent acquires diverse while effective habitual behavior?} We need stochasticity in the hidden layer ($z$) of policy network rather than stochasticity in motor actions. With proper learning process that regularizes the stochasticity, the neural network can self-develop diverse intentions, thanks to the randomness of $z$; while execute effective actions with the well-formed low-level motor skills (mapping from $z$ to motor action).

\textbf{How to bridge the gap between habitual and goal-directed behavior?} 
We propose to consider habitual and goal-directed behavior as unconditioned and goal-conditioned distribution of the latent variable $z$ (or intuitively called intention in this paper). The habitual side corresponds to the prior distribution of $z$, without conditioned on any goal, and the other side includes the hindsight goal-related information to determine the posterior of $z$. The KL-divergence term between the prior and posterior $z$, contained in the variational free energy, acts as the bridge between habitual control and goal-directed planning. Variational Bayes methods provide theoretical foundation to enable to sharing of motor skills (Sec.~\ref{chap:elbo_derive}).

\textbf{How does an agent generates actions to achieve a goal that has not been trained to accomplish?} The agent should have a internal predictive model about the environment, thus it can perform a ``mental search'' of motor patterns that results. Importantly, it is too much costly to search all possible motor actions since. Instead, intention or abstracted action ($z$ here) should be inferred -- the generated motor actions from the intention, regularized by a prior distribution (i.e., by AIf), is usually effective since well-developed low-level motor skills have already been formed. This explains why there are infinite hand gestures to hold a cup, yet we typically use the same gesture repeatedly. This share the same idea with the pre-training paradigm of the recent advances in AI such as GPT \citep{brown2020language} and CLIP \citep{radford2021learning}. In particular, designing a learning objective function for a given goal is like designing a \textit{prompt}\footnote{For language models, a prompt is a piece of text that serves as input to generate a continuation or completion, often used to guide its output.} for language models. However, a key feature of our framework is that its training process does not need to involve goals, which is different from the training of GPT where prompts are used in training.









\subsection{The frame problem of AI}

The frame problem in AI \citep{mccarthy1981some} refers to the challenge of determining which aspects of an environment are relevant when making decisions or solving problems. It arises due to the vast number of potential factors an AI system must consider, making it computationally infeasible to model all possibilities. More practically, computational models that takes the relevant information as input are the vulnerable to the frame problem.
Goal-directed behavior is more susceptible to the frame problem, as goals can be highly diverse and complex even with a single sensory modality like vision, not to mention that biological agents possess multiple sensory inputs. Rather than treating the goal as a direct input to the model \citep{andrychowicz2017hindsight}, our framework employs a backward process to infer goal-directed intentions using predictive coding. This approach presents a potential solution for the frame problem in complex environments.

\subsection{Where comes the goal}
\label{chap:discussion:where_goal}

While the proposed framework answered the question that how the goal-directed behavior using habitual skills, there remains a more fundamental question: where does the goal come from? In our simulations and many other machine learning studies \citep{andrychowicz2017hindsight, mendonca2021discovering}, the goal is assigned by the programmer per task need. But how about in humans and animals? It would be interesting to consider modelling the intrinsic mechanisms of goal-selection \citep{reinke2020intrinsically} in future research. A potential mechanism is to learn or evolve a ``meta''-habitual behavior of goal-selection that enhance the fitness of the agent and the population.

\subsection{Mutual conversion between habitual and goal-directed behaviors}
\label{chap:discussion:first_habit_or_goal}

A prevalent concept in developmental psychology is that individuals initially exhibit goal-directed behavior when adapting to new tasks or situations, which eventually evolves into more habitual and automatic responses \citep{dolan2013goals}. For example, consider a person who has just moved to a new house: initially, they need to consciously navigate using a map or directions to find their way home. Over time, as they become familiar with the route, the process of getting home transforms into a habitual and automatic behavior, requiring little conscious thought \citep{yin2006role, tricomi2009specific}. In this paper, we consider a reversed schema. Although seemingly controversial, our schema is indeed also common if readers consider the scenario that a sport player tries to win the game with certain point difference (the habitual behavior is trying to win every point). 

Nonetheless, it is straightforward to address how a repetitive goal converts to habits with our framework. The idea is similar to amortized inference \citep{gershman2014amortized}, which is a technique in machine learning that speeds up the process of inference by spreading the computational cost. Amortized inference involves training a simpler model to approximate cost-heavy inference calculations (e.g., inferring goal-directed intention), resulting in faster and more efficient predictions compared to the original inference methods. In our case, we can train a feedforward neural network $z^{\text{amortized}}(h)$ to predict the goal-directed intention from the current RNN state of the agent. The network is trained using the agent's experience during goal-directed behaviors with a repetitive goal. Our experiment shows that using $z^{\text{amortized}}(h)$ to replace the original prior and posterior intention in habitual behaving leads to fast-computing habitual behavior that only pursues the corresponding goal (See Appendix.~\ref{appendix:goal_to_habit}).

\subsection{Predictive coding and information bottleneck}
\label{chap:discussion:pc_ib}
The theory of predictive coding (PC) suggests that brain learns to identify patterns and reduce unneeded information by removing items from the input that can be predicted based on these patterns in the natural world \citep{huang2011predictive}. In our experiments, information about the environment and goal is encoded with a 2-dimensional vector $z$, while the model can still make reasonable predictions of future observations, which reflects the key idea of PC. This compact encoding together with the complexity constrain (the KL-divergence between posterior and prior), enables effective active inference to compute the optimal value of goal-directed intention with a constrained, small search space.

Another perspective from machine learning theory is that given the prior of $z$ is unit Gaussian in our cases, minimizing the expected free energy (in Equation~\ref{eq:total_loss_habitual}), or the negative variational lower bound \citep{kingma2013auto}, can be considered as a special case of the information bottleneck (IB) objective \citep{tishby2015deep, alemi2017deep}. The IB objective tends to minimize the mutual information between the input (vision here) and the latent encoding $z$ and maximize the mutual information between $z$ and the model's prediction \citep{alemi2017deep}. This idea is consistent with PC and provides a mathematical understanding of how minimizing the free energy in training relates to PC.





\subsection{Hierarchy}
\label{chap:discussion:hierarchy}
Hierarchy is a crucial property in PC because it enables efficient processing of sensory information by allowing the brain to make predictions at different levels of abstraction. E.g., neurons in the primary visual cortex have simple and complex receptive fields, while higher-level visual areas have increasingly complex receptive fields that allow for sophisticated processing of visual information, ultimately leading to object recognition \citep{rao1999predictive}. In our case, the RNN state $h$ and intention $z$ can be considered as two levels of task representation, where $z$ is of a higher-level abstraction (with only 2 dimensions compared to 256 dimensions of $h$). Nonetheless, further work should consider the intrinsic hierarchy of the model by borrowing ideas from brain science studies, such as the multiple timescale property found in cortical layers \citep{murray2014hierarchy}, or from deep learning models, such as the Swin Transformer \citep{liu2021swin}. 

\subsection{Arbitration between habitual and goal-directed behaviors}
\label{chap:discussion:arbitration}
A handful of studies consider modeling the agent's actual behavior as a mixture of goal-directed (model-based) and habitual (model-free) ones \citep{glascher2010states, smittenaar2013disruption, lee2014neural}. The most common way is to use a linear combination of actions computed by two separate systems handling goals and habits respectively. However, this is not plausible if considering the habitual behavior is going to left and the goal-directed behavior is to right in our T-maze task -- a linear combination of motor actions will lead to hitting the wall in the middle. Our Bayesian behavior framework suggests a natural way of taking advantage of both behaviors, for example, the mixed behavior can be computed using the intention $z$ given by
\begin{align}
    \label{eq:mix_z}
    z^{\text{mixed}}_t = \frac{(\sigma^p_t)^{-2} z^p_t+   (\sigma^{\text{AIf}}_t)^{-2} z^{\text{AIf}}_t}{(\sigma^p_t)^{-2} + (\sigma^{\text{AIf}}_t)^{-2}}
\end{align}
If we look at Equation~\ref{eq:mix_z}, it is easy to see that the $z_t$ (prior or goal-directed) with smaller variance ($\sigma_t^2$) is more dominant. The Bayesian property of $z$ elegantly connect the uncertainty with $\sigma_t$. Thus, the intention with lower uncertainly will be preferred. Interestingly, a state-of-the-art variational autoencoder model \citep{sonderby2016ladder, child2021very} also used similar method to compute the latent variable $z$.




\subsection{Insights for neurological diseases}

Our framework poses a novel variational Bayesian understanding about goal-directed and habitual behaviors, which may also provide valuable insights for understanding and treating neurological diseases like Parkinson's disease (PD) \citep{dolan2013goals} and autism spectrum disorder (ASD) \citep{van2014precise}.

For PD, previous research has suggested that patients with PD have difficulty with goal-directed behavior, as they tend to rely more heavily on habitual than goal-directed actions as their goal-directed planning ability is impaired \citep{wunderlich2012dopamine, wunderlich2012mapping}. As we have discussed the arbitration between the two kind of behaviors in Section~\ref{chap:discussion:arbitration}, such impairment may be explained by large uncertainty of goal-directed intention. It might be worth investigating how to reduce the uncertainty of goal-directed intention through medicine / deep brain stimulation (changing internal states) \citep{cotzias1969modification, perlmutter2006deep} or sensory stimulus (changing brain inputs) \citep{azulay1999visual, munoz2013visual} for improving the motor control ability of PD patients.

It is well known that repetitive behavior is a key characteristic of ASD and the abnormal predictive coding in ASD people is a popular explanation \citep{pellicano2012world, wild2012goal, van2014precise, palmer2017bayesian}. In particular, the pathology of ASD can be computationally explained by that over-weighting of the complexity term in free energy (Equation~\ref{eq:free_energy_abstract}) impairs cognitive-behavioral flexibility when adapting changing environment \citep{soda14simulating}. Our framework can be used as a computational tool to help understand how the stochasticity of $z$ affects behavioral diversity (Appendix~\ref{appendix:sweep}).

\section{Conclusion}

In this article, we proposed the Bayesian behavior framework, suggesting a novel paradigm to consider habitual and goal-directed behavior. Our contributions are two-fold -- technical and conceptual.

Technically, we propose a novel Bayesian framework that enables seamless transfer between fast, inflexible habitual behavior and slow, flexible goal-directed behavior. The proposed Bayesian behavior framework is based on two core ideas -- modeling habits and goals with prior and posterior distributions of a Bayesian variable, respectively; and utilizing the predictive internal model for goal-directed planning. Based on these two idea, we particularly employed model-free deep RL to learn motor skills for habitual behavior, since deep RL catches many key features of embodied learning of biological agents \citep{botvinick2020deep}. We then apply AIf which enables flexible goal-directed behavior \citep{friston2010action}. Despite the common belief that AIf and RL are contradictory \citep{friston2009reinforcement}, we argue that combining the two methods results in more efficient learning and reuse of motor skills. Our framework provides a concrete methodology that elegantly unit these all together.

For the habitual behavior, we have shown the emergence of smooth bifurcating behavior of an embodied AI through trial and error (i.e. online deep RL) using the proposed framework (Section~\ref{chap:habitual_results}). Simultaneously, the agent learns an internal model to predict future observations conditioned on the intention $z$. Then, we have demonstrated the flexibility of goal-directed behavior with the framework by leveraging the predictive internal model (Section~\ref{chap:gdp_results}). Although having not being trained with goals, the agent is able to perform planning for a given goal observation or partial goal properties.



Conceptually, our work proposes potential underlying neural mechanisms that are essential to flexible and efficient behavior of intelligence biological agents. To briefly summarize the take-away messages of this work for cognitive science and neuroscience researchers:
\begin{itemize}
    \item The stochasticity in neural activities with proper regularization enables diverse and effective motor actions.
    \item Habitual and goal-directed behavior, though having different conditions, can share low-level motor skills using variational Bayesian methods.
    \item Predictive coding enables flexible goal-directed behavior by inferring what values of neural state $z$ result in the desired goal properties.
    \item The variational free energy's complexity term bridges the gap between habitual and goal-directed behaviors.
\end{itemize}

Our framework has certain limitations. We mainly focus on proof of concept using a fundamental T-maze task, which is nevertheless challenging due to the high-dimensional nature of the first-person vision used for observation. We have yet to tackle more complex motor control tasks, which are commonly seen in real animal behavior \citep{mattar2022planning}.

Another limitation is that the agent may not be able to reach a state or place that is not covered by its habitual behavior. This scenario is relatively rare, but it may occur (e.g., an intentional loss to a weaker opponent in a sport game). To overcome this limitation, it may be necessary to conduct additional learning or to search raw actions rather than relying on the learned motor skills, which can be more time-consuming and resource-intensive.

As for future research, intrinsic generation of goals (Sec.~\ref{chap:discussion:where_goal}) appears an important direction to answer the ultimate questions of how autonomous agents can self-develop \citep{lecun2022path}. Another important change lies on hierarchy model structures (Section~\ref{chap:discussion:hierarchy}). Moreover, significant improvement may come from integrating more modalities such as nature language and sound. The format of the goal could be much more flexible if introducing pretrained models like GPT \citep{brown2020language} and CLIP \citep{radford2021learning}. These models provides extensive possibilities of incorporating different modalities into goal-directed intention.

\section*{Acknowledgement}
This work was supported by Okinawa Institute of Science and Technology and Microsoft. Kenji Doya was also supported by Japan Society for the Promotion of Science KAKENHI Grant Numbers JP16K21738, JP16H06561 and JP16H06563. The authors would like to thank the members in cognitive neurorobotics research unit, neural computation unit of OIST, and AI/ML group in MSRA shanghai; as well as Zhaoyun Chen for insightful discussions and comments.


\bibliography{paper}
\bibliographystyle{icml2022}

\newpage
\appendix
\onecolumn
\section{Appendix}

\subsection{Convolutional and de-convolutional networks structure}

The structures of the convolutional and de-convolutional neural networks used for image encoding and decoding in this work are specified in Table~\ref{table:cnn_structure} and Table~\ref{table:dcnn_structure}.

\begin{table}[h]
\centering
\begin{tabular}{|c|c|c|c|c|c|c|}
\hline
Layer & Type & Kernel Size & Stride & Padding & Channels & Activation \\
\hline
1 & Conv2d & $(4, 4)$ & $(2, 2)$ & $(1, 1)$ & 8 & ReLU \\
\hline
2 & Conv2d & $(4, 4)$ & $(2, 2)$ & $(1, 1)$ & 16 & ReLU \\
\hline
3 & Conv2d & $(4, 4)$ & $(2, 2)$ & $(1, 1)$ & 16 & ReLU \\
\hline
4 & Conv2d & $(2, 4)$ & $(2, 2)$ & $(0, 1)$ & 64 & ReLU \\
\hline
5 & Conv2d & $(1, 4)$ & $(1, 4)$ & $(0, 0)$ & 256 & ReLU \\
\hline
6 & Flatten & - & - & - & - & -\\
\hline
\end{tabular}
\caption{Structure of the convolutional neural network.}
\label{table:cnn_structure}
\end{table}

\begin{table}[h]
\centering
\begin{tabular}{|c|c|c|c|c|c|c|}
\hline
Layer & Type & Kernel Size &  Stride &  Padding & Channels & Activation \\
\hline
1 & ConvTranspose2d & (1, 4) &  (1, 1) &  (0, 0) & 64 & ReLU \\
\hline
2 & ConvTranspose2d & (2, 4) &  (1, 2) &  (0, 1) & 16 & ReLU \\
\hline
3 & ConvTranspose2d & (4, 4) &  (2,2) &  (1, 1) & 16 & ReLU \\
\hline
4 & ConvTranspose2d & (4, 4) &  (2, 2) &  (1, 1) & 8 & ReLU \\
\hline
5 & ConvTranspose2d & (3, 3) &  (2, 2) &  (1, 1) & 8  & ReLU \\
\hline
6 & Conv2d & (3, 3) &  (1, 1) &  (1, 1) & 3 & - \\
\hline
\end{tabular}
\caption{Structure of the de-convolutional neural network.}
\label{table:dcnn_structure}
\end{table}

\subsection{Repetitive goal converts to habit}
\label{appendix:goal_to_habit}
\begin{figure}[h]
    \centering
    \includegraphics[width=0.5\textwidth]{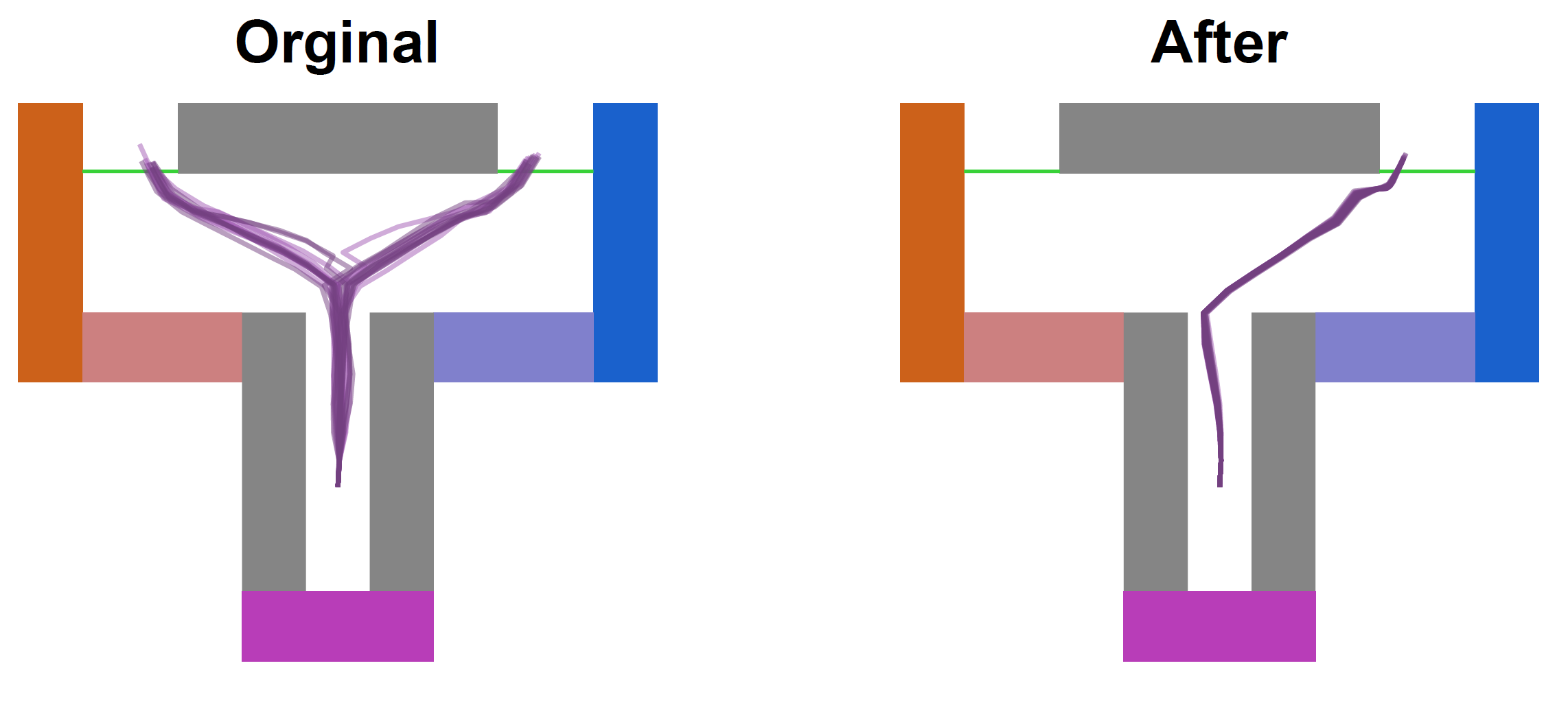}
    \caption{An agent's original habitual behavior in the T-maze (left), and its behavior after amortized inference of $z$ with 10 trials of goal-directed task in which the goal observation is from the top-right exit (right). 30 trials with deterministic motor actions are plotted in each case.}
    \label{fig:goal2habit}
\end{figure}

If a goal is repetitively occurring, human and animals will eventually turn the slow goal-directed planning into fast habitual control \citep{dolan2013goals}. We here demonstrate that this phenomena can be explained by the agent's amortized inference of $z$ (Section~\ref{chap:discussion:first_habit_or_goal}). For detailed implementation, a fully-connected feedforward network $z^{\text{amortized}}(h)$ (with 1 hidden layer of 256 neurons and ReLU activation) is trained to estimate $z^{\text{AIf}}_t$ from $h^q_{t-1}$. The data of $z^{\text{AIf}}_t$ and $h^q_{t-1}$ are obtained from 10 trials of goal-directed planning (Section~\ref{chap:methods:gdp}) with the same goal (providing goal observation at the top-right exit of T-maze). We also add a diagonal Gaussian white noise $\mathcal{N}(\bm{0}, 0.1I)$ to $z^{\text{amortized}}$ to hold its stochasticity.

We use an Adam optimizer with learning rate 0.0003 (same as the optimizer used in training the model) to train $z^{\text{amortized}}(h)$ for 1,000 epochs. Then, the prior and posterior $z_t$ are replaced with $z^{\text{amortized}}(h^q_{t-1})$ to compute the new habitual behavior, which always chooses to exit from top-right of the maze (Figure~\ref{fig:goal2habit} right).

\subsection{Weighting of the complexity term}
\label{appendix:sweep}
The weighting of the complexity term ($\beta_z$ in Equation~\ref{eq:total_loss_habitual},\ref{eq:aif_loss}) affects the trade-off between accuracy and complexity (Section~\ref{chap:discussion:pc_ib}). Roughly speaking, higher $\beta_z$ leads to low complexity (less information in $z^q$) and less accurate prediction, and vice versa \citep{higgins2017beta}. We sweep a range of $\beta_z$ (1, 10, 100 (used in the paper), 1,000 and 10,000) to check the impact of it to the learned habitual behavior.  

\begin{figure}
    \centering
    \includegraphics[width=0.72\textwidth]{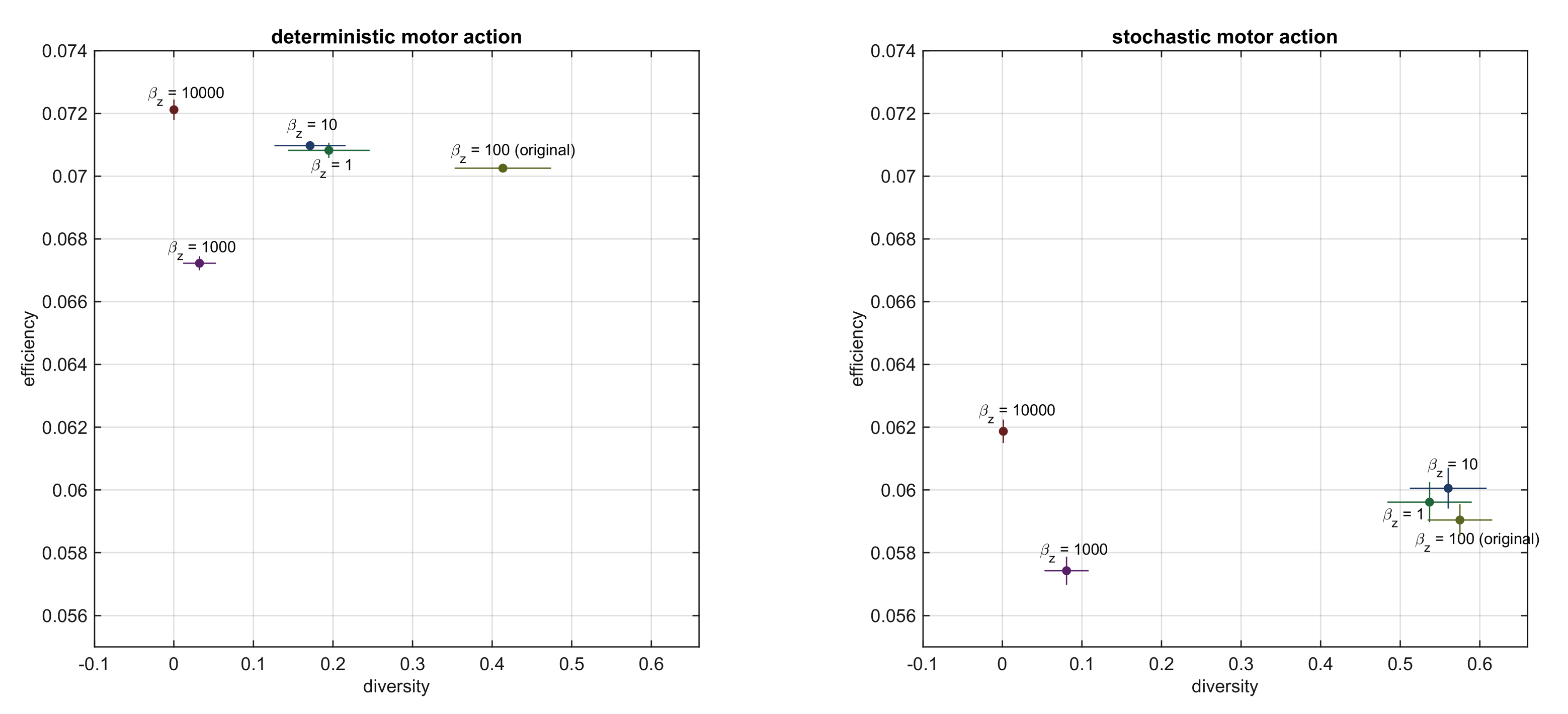}
    \caption{Diversity and efficiency of the learned habitual behavior using different $\beta_z$, plotted in the same way as Figure~\ref{fig:exit_ratio}a,c.}
    \label{fig:sweep}
\end{figure}

It can be seen that (Figure~\ref{fig:sweep}) agents with $\beta_z=100$ overall have the best overall behavior efficiency and diversity. In particular, when $\beta_z$ is very large, the agent tend to acquire habitual behavior with very low diversity. 



\end{document}